\documentclass[11pt]{article}

\usepackage[preprint]{acl}

\usepackage{times}
\usepackage{latexsym}
\usepackage[T1]{fontenc}
\usepackage[utf8]{inputenc}
\usepackage{microtype}
\usepackage{inconsolata}
\usepackage{graphicx}
\usepackage{amsmath}
\usepackage{booktabs}
\usepackage{xcolor}
\usepackage{xspace}

\newcommand{\nll}{\mathrm{NLL}}
\newcommand{\cv}{\mathrm{CV}}
\newcommand{\tvs}{\ensuremath{\mathrm{TVS}}\xspace}

\title{What Improves During Pretraining? \\ Tracking Knowledge Acquisition with Point-in-Time Models}

\author{Chang Su$^{\alpha}$ \quad
Bin Wu$^{\beta}$ \quad
Emine Yilmaz$^{\beta}$ \quad
Yao Lu$^{\beta}$ \\
$^{\alpha}$University of Pennsylvania \\
$^{\beta}$Centre for Artificial Intelligence, University College London \\
\texttt{sucha@engineering.upenn.edu}\quad
\texttt{bin.wu.23@ucl.ac.uk} 
\\
\texttt{emine.yilmaz@ucl.ac.uk} \quad
\texttt{yao.lu@cs.ucl.ac.uk}
}

\begin{document}
\maketitle

\begin{abstract}

Language models pretrained on web corpora collected over multiple years learn both general and domain-specific knowledge. How these different kinds of knowledge are learned as pretraining progresses remains unclear. Recent point-in-time models trained on web snapshots from different years provide a way to probe how knowledge prevalence varies over time. Using these models, we propose a temporal variability score (TVS), defined as the coefficient of variation of an item's negative log-likelihood across models. A higher score suggests that the knowledge may be less consistently represented across years of web text. We evaluate six benchmarks covering diverse topics using six models from multiple open-source model families. We find that low- and high-TVS items are learned similarly early in training, but high-TVS items improve faster later on. At the end of the pretraining run, high-TVS items outperform low-TVS items in 34 of 36 model--benchmark settings, with an average advantage of 7.3 points of chance-normalized accuracy. This pattern is most robust on knowledge-intensive tasks, even when comparing items of similar difficulty. Our findings reveal differences in how knowledge is learned over the course of pretraining, with temporal variability providing a diagnostic of these learning dynamics beyond aggregate benchmark scores.

\end{abstract}

\section{Introduction}

Language models improve on a wide range of tasks as pretraining progresses, yet these gains reveal little about the characteristics of the knowledge being learned \citep{swayamdipta-etal-2020-dataset, chang-etal-2024-factual}. We ask how different kinds of knowledge are learned at different stages of pretraining, using their prevalence in training data as a starting point for characterizing these differences.

However, measuring how often a fact or concept appears in pretraining data is difficult because the same knowledge can be expressed in many different ways. Corpus search tools such as infini-gram \citep{xu-etal-2025-infinigram-mini} count exact textual matches, but these counts depend on the wording of the query. Passages that express the same knowledge in different words may go uncounted, making exact-match frequency an incomplete measure of prevalence.

The temporal structure of web corpora offers a complementary approach. Pretraining data combine web crawls collected in different years. Our intuition is that knowledge that appears consistently across these years should receive similar likelihoods from models trained on each yearly snapshot, even when its wording differs. Knowledge whose prevalence varies across years may instead receive more variable likelihoods. Point-in-time (PIT) language models, which share an architecture and training recipe but are trained on distinct yearly web snapshots \citep{drinkall-etal-2024-time,he2025chronogpt,yan-etal-2026-datedgpt,kelly-etal-2026-pit}, allow us to probe this variation. We propose a temporal variability score (\tvs), defined as the coefficient of variation of the negative log-likelihood assigned to an item's question and correct answer across eight yearly PIT-4B models \citep{kelly-etal-2026-pit}. Higher scores suggest that the underlying knowledge appears less consistently across years of web text. \tvs measures likelihood variation, not prevalence directly.

\tvs enables a fine-grained analysis of knowledge acquisition during pretraining. We use individual items from widely used benchmarks spanning factual knowledge, scientific reasoning, and commonsense reasoning to assess what models learn at different stages. We compute \tvs for 16,239 items from six multiple-choice benchmarks, divide each benchmark into length-matched low-, medium-, and high-\tvs groups, and track their accuracy across checkpoints of six models at different scales. We examine whether low- and high-\tvs items exhibit different learning patterns at different stages of training.

We find that low- and high-\tvs items perform similarly early in training, but high-\tvs items improve faster at later stages. At the end of the pretraining runs, high-\tvs items outperform low-\tvs items in 34 of 36 model--benchmark settings, with an average advantage of 7.3 points of chance-normalized accuracy. The pattern is most robust on knowledge-intensive tasks: on ARC and MMLU, the advantage remains positive in all 18 settings when comparing items of similar difficulty. These findings connect temporal variability to differences in knowledge acquisition that aggregate benchmark scores can mask.

Our contributions are as follows:
\begin{itemize}
    \item We propose \tvs, an item-level diagnostic that uses point-in-time models to characterize sensitivity to training-data composition beyond exact textual matches.
    \item We document differences in learning across \tvs groups: items with higher temporal variability improve more at later training stages, with the most robust evidence on knowledge-intensive tasks.
    \item We show that, on knowledge-intensive tasks, the component of \tvs specific to the correct answer predicts growth in models' preference for that answer during later training stages, among similarly difficult items.
\end{itemize}

\section{Measuring Temporal Variability}

\paragraph{Point-in-time models.}
We use PIT-4B \citep{kelly-etal-2026-pit}, a suite of 4B-parameter base models trained on yearly partitions of FineWeb \citep{penedo-etal-2024-fineweb}. The eight models for 2017--2024 share architecture, tokenizer, and training setup and differ in the crawl year of their training data. Crawl years index when pages were collected, not when facts originated: each snapshot also contains older, long-lived content. Related PIT and time-stratified models are discussed in \S\ref{sec:related}.

\paragraph{Temporal variability score (TVS).}
We define the temporal variability score using the coefficient of variation (CV) across training snapshots: for item $i$, let $s_i$ be the string formed by the question and its gold answer, and let $\nll_y(s_i)$ be the per-token negative log-likelihood of $s_i$ under the PIT model for year $y \in \{2017,\ldots,2024\}$. We define
\begin{equation}
  \mathrm{TVS}(i) =
  \frac{\operatorname{std}_{y}(\nll_y(s_i))}
  {\operatorname{mean}_{y}(\nll_y(s_i))}.
\end{equation}
Low \tvs means that the yearly models agree on the item relative to its overall likelihood; we call such items \emph{temporally stable}. High \tvs means that the item's likelihood depends on the training year; we call such items \emph{temporally variable}. Normalizing by the mean matters. The coefficient of variation of perplexity is strongly entangled with mean perplexity; unnormalized NLL standard deviation mostly selects harder items (Appendix~\ref{app:confounds}).

\paragraph{Interpreting TVS.}
\tvs measures relative cross-year variation in likelihood, not prevalence directly or semantic categories such as ``commonsense'' versus ``specialized.'' Each yearly model scores the same fixed question--answer string: time indexes the training snapshot, not a changing evaluation target. Stable and variable are likelihood-based labels; Table~\ref{tab:examples} illustrates groups, not a knowledge taxonomy. We also isolate correct-answer variation from the shared question and format (\S\ref{sec:where}; Appendix~\ref{app:item-level}).

\begin{table*}[t]
  \centering
  \small
  \resizebox{0.95\textwidth}{!}{\begin{tabular}{p{0.12\linewidth} p{0.12\linewidth} p{0.68\linewidth}}
    \toprule
    Benchmark & TVS group & Example (question $\rightarrow$ gold answer) \\
    \midrule
    PIQA & Low & ``How do you take soap off of clean dishes?'' $\rightarrow$ ``rinse it in water.'' \\
    PIQA & High & ``To tell when baked goods are done baking, you can'' $\rightarrow$ ``Wait for them to turn golden brown.'' \\
    ARC-Easy & Low & ``What causes a change in the speed of a moving object?'' $\rightarrow$ ``Force.'' \\
    ARC-Easy & High & ``A meteoroid becomes a meteorite when it'' $\rightarrow$ ``impacts the surface of a planet.'' \\
    \bottomrule
  \end{tabular}}
  \caption{Sampled items from the low- and high-\tvs groups.}
  \label{tab:examples}
\end{table*}

\section{Experimental Setup}

\paragraph{Datasets.}
We evaluate on HellaSwag \citep{zellers-etal-2019-hellaswag}, PIQA \citep{bisk-etal-2020-piqa}, WinoGrande \citep{sakaguchi-etal-2020-winogrande}, ARC-Easy and ARC-Challenge \citep{clark-etal-2018-arc}, and MMLU \citep{hendrycks-etal-2021-mmlu}. For each benchmark we sample 3,000 items from the training and validation splits (the test split for MMLU), or use all items when fewer are available, for 16,239 items in total. \tvs is correlated with string length, so we stratify within length. We divide each benchmark into ten bins by PIT-4B token length, split each bin into equal low-, medium-, and high-TVS thirds, and pool the bins (Appendix~\ref{app:length-control}).

\paragraph{Models.}
We evaluate checkpoints of six models from four open families\footnote{Checkpoints cover most of each training run, from 0.54B to 11.1T tokens; each trajectory is a single run.}: Pythia 1.4B and 6.9B \citep{biderman-etal-2023-pythia}, TinyLlama 1.1B \citep{zhang-etal-2024-tinyllama}, SmolLM3 3B \citep{huggingface-2025-smollm3}, and OLMo 1B and 7B \citep{groeneveld-etal-2024-olmo}.\footnote{We use the 0724 versions of both OLMo models.}

\paragraph{Evaluation metric.}
Each item is scored by ranking its answer options by length-normalized likelihood. We report chance-normalized accuracy:
\begin{equation}
  \operatorname{acc}_{\text{norm}} =
  \frac{\operatorname{acc}_{\text{raw}} - \operatorname{chance}}
       {1 - \operatorname{chance}}.
\end{equation}
This maps random guessing to 0 and perfect accuracy to 1 and makes benchmarks with different numbers of options comparable \citep{yan-etal-2026-datedgpt,kelly-etal-2026-pit}.

\begin{figure*}[t]
  \centering
  \includegraphics[width=\textwidth,height=.72\textheight,keepaspectratio]{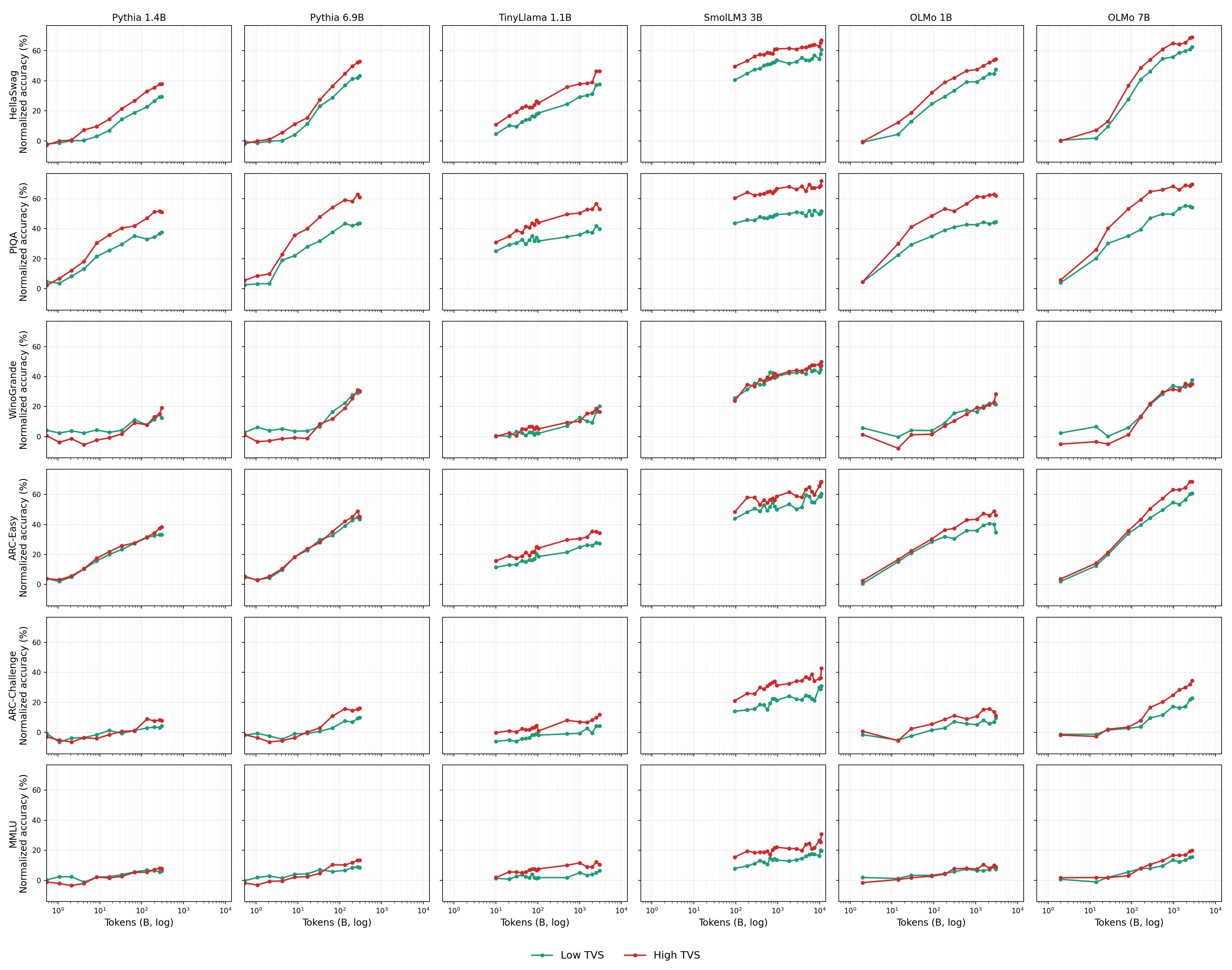}
  \caption{Chance-normalized accuracy over pretraining for temporally variable (high \tvs, red) and temporally stable (low \tvs, green) items, by model (columns) and benchmark (rows). Groups are length-matched; training tokens are on a log scale. Pythia and OLMo show little early separation; high-\tvs items generally pull ahead later.}
  \label{fig:full-grid}
\end{figure*}

\section{Results}

\subsection{Learning Patterns Diverge Across TVS Groups}

We examine how learning patterns differ across low- and high-\tvs items as training progresses (Figure~\ref{fig:full-grid}). Our earliest Pythia and OLMo checkpoints occur after 0.54B and 2B training tokens, respectively. At these checkpoints, low-\tvs items outperform high-\tvs items in 13 of 24 settings, no more often than expected by chance. The mean high-minus-low gap is $-0.8$ points of chance-normalized accuracy. The groups separate later: over the second half of the checkpoints, accuracy on high-\tvs items grows faster by 2.0 points per tenfold increase in training tokens. The difference is positive in 26 of 36 settings and in all six benchmark and model averages, but not significant across four family averages ($p=.13$; Appendix~\ref{app:effect-sizes}).

At the end of the pretraining runs, high-\tvs items outperform low-\tvs items in 34 of 36 model--benchmark settings (Table~\ref{tab:summary}). The mean gap is 7.3 points of chance-normalized accuracy (95\% CI [5.5, 9.2]). It remains significant when settings are first averaged within each of the four model families (CI [4.8, 10.2], $p=.003$), and it is positive on every benchmark, ranging from 1.7 points on WinoGrande to 16.1 points on PIQA (Appendix~\ref{app:effect-sizes}). These intervals summarize variation across settings or family averages, not run-to-run uncertainty: each setting contains one training run. The pattern holds across model families trained on different corpora, which suggests that it is not tied to a single training recipe.

\begin{table}[t]
  \centering
  \footnotesize
  \setlength{\tabcolsep}{3pt}
  \begin{tabular}{@{}lcccc@{}}
    \toprule
    & Early & \multicolumn{3}{c}{Final} \\
    \cmidrule(lr){2-2}\cmidrule(lr){3-5}
    Benchmarks & L$>$H & H$>$L & Gap & Gap $\mid$ NLL \\
    \midrule
    All & 13/24 & 34/36 & +7.3 & +3.7 \\
    HS/PIQA/WG & 8/12 & 16/18 & +8.5 & +0.5 \\
    ARC/MMLU & 5/12 & 18/18 & +6.2 & +6.9 \\
    \bottomrule
  \end{tabular}
  \caption{Low- (L) versus high-TVS (H) items at the earliest and final checkpoints. Early counts cover Pythia and OLMo, whose first checkpoints come early in training; final counts cover all six models. Gaps are high-minus-low chance-normalized accuracy in points. Gap $\mid$ NLL uses a split conditioned on mean NLL (\S\ref{sec:difficulty}).}
  \label{tab:summary}
\end{table}

\subsection{Does TVS Capture More Than Question Difficulty?}
\label{sec:difficulty}

To distinguish the relationship between \tvs and learning from differences in question difficulty, we repeat the analysis using the variation in \tvs not explained by mean NLL. We regress \tvs on linear and quadratic mean-NLL terms within each benchmark and length bin, then divide the residuals into thirds. Within each length bin, high-\tvs items have slightly lower mean NLL, making this adjustment relevant to the comparison. On ARC and MMLU, the advantage remains positive in all 18 settings, at +6.9 points of chance-normalized accuracy (Table~\ref{tab:summary}). On HellaSwag, PIQA, and WinoGrande, the smaller gap (+0.5 points) indicates a substantial difficulty contribution (Appendix~\ref{app:mean-control}).

Conditioning additionally on PIT-4B's gold-versus-distractor margin and accuracy leaves a final gap of 1.4 points, positive in 22 of 36 settings (Table~\ref{tab:splits}). These covariates capture average likelihood and answer discrimination in the diagnostic suite; they need not exhaust difficulty for every evaluated model. The raw 7.3-point gap is not an estimate of an independent \tvs effect; the remaining association varies across benchmarks.

We also examine how the results depend on the measure used to group the same items (Table~\ref{tab:splits}). Splitting directly by mean NLL produces the opposite ordering: the higher-NLL group performs worse in all 36 final settings. Replacing \tvs with the unnormalized standard deviation of NLL also reverses the effect. This measure increases with mean NLL and therefore selects harder items. These comparisons distinguish the \tvs pattern from those obtained by grouping items by mean NLL or unnormalized dispersion.

\begin{table}[t]
  \centering
  \footnotesize
  \setlength{\tabcolsep}{2.5pt}
  \begin{tabular}{@{}lcc@{}}
    \toprule
    Split (within length bins) & Final H$>$L & Gap \\
    \midrule
    \tvs (main) & 34/36 & +7.3 \\
    \tvs $\mid$ mean NLL & 26/36 & +3.7 \\
    \tvs $\mid$ mean NLL, PIT margin, PIT acc. & 22/36 & +1.4 \\
    Gold-vs.-wrong \tvs margin & 36/36 & +12.3 \\
    Question-only \tvs & 26/36 & +2.1 \\
    \midrule
    NLL standard deviation & 8/36 & $-$2.6 \\
    Mean NLL quintiles (high = harder) & 0/36 & $-$31.1 \\
    \bottomrule
  \end{tabular}
  \caption{Final-checkpoint comparison of high- versus low-group items under alternative stratifications (chance-normalized points; 36 model--benchmark settings). Conditioned splits use the residual of \tvs after regressing out the listed covariates within each benchmark and length bin. Full results are in Table~\ref{tab:app-split-summary}.}
  \label{tab:splits}
\end{table}

\subsection{Answer-Specific TVS Predicts Learning in Later Training Stages}
\label{sec:where}

Answer options share a question and format, so we isolate correct-answer variation as its \tvs minus the mean \tvs of incorrect answers. We test whether it predicts growth in the correct-answer-versus-best-distractor margin, measured as its slope against log training tokens over the upper half of the log-token range. Margin growth need not change the top-ranked option, so we analyze it separately from final accuracy. Regressions include shared and answer-specific components, controlling for PIT-4B mean NLL, margin, and accuracy. The answer-specific signal predicts greater late margin growth on ARC-Challenge, ARC-Easy, and MMLU ($p<10^{-6}$ each; Table~\ref{tab:app-item-level}), with a weaker effect on PIQA. On HellaSwag and WinoGrande, it predicts a small final-accuracy difference but no robust late-margin growth (Appendix~\ref{app:item-level}).

Group comparisons agree: the correct-answer-versus-strongest-distractor \tvs margin yields the largest final gap, with the high group ahead in all 36 settings by 12.3 points on average and the best final group in 35. In contrast, question-only \tvs yields an average gap of just 2.1 points (Appendices~\ref{app:gold-wrong} and~\ref{app:question-only}). Thus the relationship depends substantially on the answer itself, rather than only on question wording.

\begin{table}[t]
  \centering
  \footnotesize
  \begin{tabular}{@{}lcc@{}}
    \toprule
    Benchmark & Final correct & Late margin slope \\
    \midrule
    ARC-C & +0.024 (.04) & +0.070 ($2\times10^{-9}$) \\
    ARC-E & +0.008 (.2) & +0.059 ($2\times10^{-8}$) \\
    MMLU & +0.001 (.8) & +0.043 ($2\times10^{-7}$) \\
    PIQA & +0.001 (.8) & +0.011 ($2\times10^{-4}$) \\
    HellaSwag & +0.019 ($10^{-5}$) & 0.000 (.8) \\
    WinoGrande & +0.012 (.01) & +0.002 (.08) \\
    \bottomrule
  \end{tabular}
  \caption{Difficulty-controlled effect of gold-specific \tvs (per standard deviation) on final correctness and late gold-versus-distractor margin growth ($p$-values in parentheses; Appendix~\ref{app:item-level}).}
  \label{tab:app-item-level}
\end{table}

\subsection{Assessing TVS Robustness to Benchmark Contamination}
\label{sec:contamination}

We test for benchmark exposure in PIT-4B using release dates and textual overlap (Appendix~\ref{app:contamination}). Benchmark copies should raise gold-answer likelihood after release. Yet staggered difference-in-differences with item and year fixed effects finds no decrease in gold-answer NLL ($+0.016$) or detectable change in the gold-versus-distractor margin ($p=.34$). High-\tvs accuracy falls slightly ($-0.011$, $p=.004$), opposite the expected leakage effect. ARC, PIQA, and MMLU share similar year profiles despite different release dates, with no release-specific gains.

We also scan 55B FineWeb tokens, using ten files per crawl year, for 13-gram overlap with each item \citep{brown-etal-2020-language}. Overlap is rare (3.1\% of high-\tvs versus 2.0\% of low-\tvs items) and mostly comes from source pages such as WikiHow rather than answer keys. Years with more overlap do not have lower PIT-4B NLL for the matched items. Adding overlap indicators to the item-level regressions leaves the \tvs coefficient essentially unchanged. These checks provide evidence against benchmark leakage as the dominant explanation of the \tvs pattern in PIT-4B, but sparse yearly sampling cannot establish its absence or exclude contamination of the evaluated models' corpora.

\section{Related Work}
\label{sec:related}

\paragraph{Benchmark analysis.}
Dataset cartography characterizes items through a fine-tuned model's training dynamics \citep{swayamdipta-etal-2020-dataset}, while item response theory estimates their difficulty and discrimination \citep{lalor-etal-2016-building,martinez-plumed-etal-2019-irt}. Other approaches group benchmark items by difficulty or composition \citep{ding-etal-2024-easy2hard,kovatchev-lease-2024-benchmark}. These measures describe items through the behavior of the evaluated models and can therefore vary with the model being diagnosed. \tvs instead uses an external suite that shares architecture and training setup but differs in its temporal data slice. It therefore characterizes items independently of the evaluated model's own training trajectory and supports comparisons across model families.

\paragraph{Attributing learning to pretraining data.}
Models learn facts through repeated exposure \citep{chang-etal-2024-factual}, memorize more over training \citep{tirumala-etal-2022-memorization}, and struggle with rare knowledge \citep{kandpal-etal-2023-longtail}. Corpus search counts exact matches \citep{elazar-etal-2024-wimbd,xu-etal-2025-infinigram-mini}, making exposure estimates wording-dependent. Controlled ablations link capabilities to corpus components: removing parallel bilingual documents impairs translation but largely preserves cross-lingual question answering and reasoning \citep{shao-etal-2026-role}. \tvs complements these approaches by measuring sensitivity beyond exact overlap without requiring corpus interventions.

\paragraph{Point-in-time language models.}
Time-stratified models probe temporal generalization and time-sensitive facts \citep{lazaridou-etal-2021-mind,dhingra-etal-2022-time}; PIT suites also address lookahead bias and historically bounded knowledge \citep{drinkall-etal-2024-time,he2025chronogpt,yan-etal-2026-datedgpt,kelly-etal-2026-pit}. We instead characterize fixed benchmark items through yearly likelihood variation, defining groups independently of evaluated checkpoints and comparing trajectories without their training corpora. Time indexes the corpus, not an intervention or changing targets.

\section{Conclusion}

We introduced \tvs to track item-level learning using point-in-time models. Across six benchmarks and six models, high-\tvs items generally improve more later; the earliest Pythia and OLMo checkpoints show little group separation. Effects are most robust on ARC and MMLU, where difficulty-controlled, answer-specific \tvs predicts late margin growth. \tvs reveals dynamics beyond aggregate scores without requiring training-corpus access or assigning semantic categories to items.

\section*{Limitations}

\tvs is a model-based likelihood statistic, and its effectiveness therefore depends on the quality of the underlying model(s). The PIT-4B models cover crawl years 2017--2024 at a single model size. 
Our contamination checks cover the PIT-4B training data rather than the full pretraining corpora of the evaluated models (Appendix~\ref{app:contamination}). Finally, we study only likelihood-ranked multiple-choice benchmarks as they are most suitable when monitoring models' early checkpoints.

\section*{Acknowledgments}

This research is supported by the Engineering and Physical Sciences Research Council [grant number UKRI5028] and a 2025 Amazon Research Award. We also gratefully acknowledge NVIDIA Corporation for their support through the NVIDIA Academic Grant Program, which provided GPU computing resources used in this research. Part of this research was supported by the Bloomberg Data Science PhD Fellowship programme.

\bibliography{custom}

\clearpage
\appendix

\section{Additional Results}
\label{app:full-dynamics}

\begin{figure*}[t]
  \centering
  \includegraphics[width=\textwidth,height=.74\textheight,keepaspectratio]{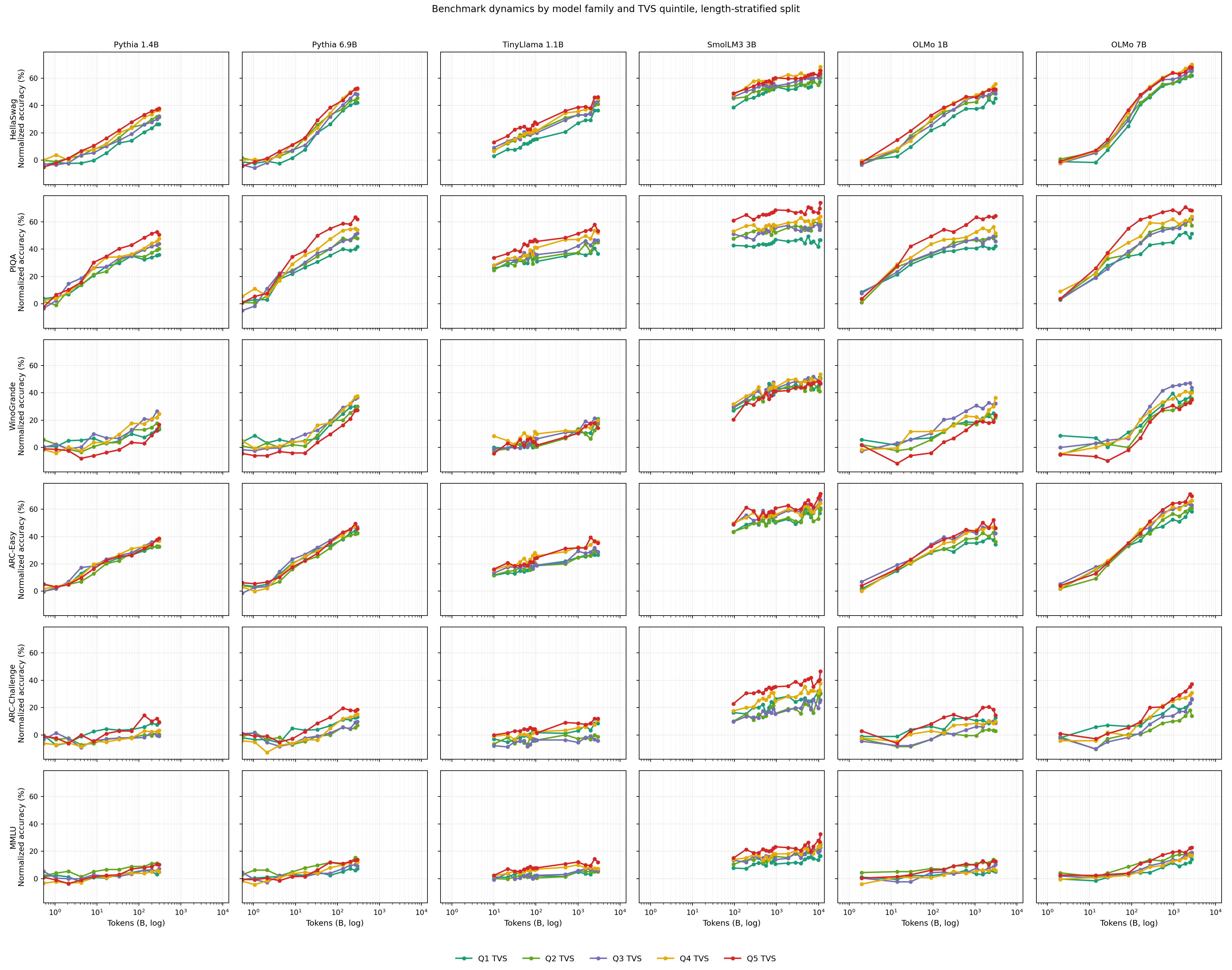}
  \caption{Main result with five \tvs quantiles. The most variable quintile ends ahead of the least variable one in 32 of 36 settings; all four exceptions are on WinoGrande.}
  \label{fig:app-gold-q5}
\end{figure*}

In the appendix, gaps are reported as fractions of chance-normalized accuracy (e.g., 0.073 corresponds to 7.3 points in the main text). Table~\ref{tab:app-split-summary} compares the main split with alternative stratifications. For each model--benchmark setting we compare the earliest and final checkpoints. ``Early L$>$H'' counts settings in which the low group is ahead at the earliest checkpoint; unlike Table~\ref{tab:summary}, it includes all six models. ``Final H$>$L'' counts the reverse at the final checkpoint. ``Best H'' counts settings in which the high group has the highest final accuracy of all groups. No setting had exactly tied accuracies. Figure~\ref{fig:app-gold-q5} shows the main result with five \tvs quantiles instead of three.

\begin{table*}[!t]
  \centering
  \footnotesize
  \begin{tabular}{@{}lcccc@{}}
    \toprule
    Stratification & Early L$>$H & Final H$>$L & Best H & Mean final H$-$L \\
    \midrule
    Gold TVS tertiles (main) & 15/36 & 34/36 & 28/36 & +0.073 \\
    Gold TVS quintiles & 15/36 & 32/36 & 25/36 & +0.081 \\
    Mean gold NLL quintiles & 35/36 & 0/36 & 0/36 & $-$0.311 \\
    Question-only TVS tertiles & 13/36 & 26/36 & 20/36 & +0.021 \\
    Gold-vs.-wrong TVS margin tertiles & 20/36 & 36/36 & 35/36 & +0.123 \\
    \midrule
    Gold NLL-SD tertiles & 30/36 & 8/36 & 7/36 & $-$0.026 \\
    Rank-disagreement tertiles & 15/36 & 22/36 & 16/36 & +0.011 \\
    TVS $\mid$ mean NLL tertiles & 18/36 & 26/36 & 22/36 & +0.037 \\
    TVS $\mid$ mean NLL, PIT margin, PIT acc.\ tertiles & 21/36 & 22/36 & 18/36 & +0.014 \\
    \bottomrule
  \end{tabular}
  \caption{Alternative stratifications across 36 model--benchmark settings. Mean final H$-$L is the final-checkpoint high-minus-low gap in chance-normalized accuracy. For mean gold NLL, ``high'' is the higher-NLL group; for the gold-vs.-wrong margin, ``high'' has the larger $\mathrm{TVS}(q,a^{\mathrm{gold}})-\mathrm{TVS}(q,a^{\mathrm{wrong}})$. All splits are formed within ten length bins per benchmark; the question-only split bins by question length. NLL-SD is the unnormalized standard deviation of gold NLL across years. Rank disagreement is the standard deviation across years of an item's NLL percentile within its benchmark and length bin. The conditioned rows split on the residual of \tvs after regressing it (with quadratic terms) on the listed covariates within each benchmark and length bin (Appendix~\ref{app:mean-control}).}
  \label{tab:app-split-summary}
\end{table*}

\subsection{Effect Sizes and Significance}
\label{app:effect-sizes}

All values are in chance-normalized accuracy, and each model--benchmark setting has one training run. The mean final high-minus-low gap across the 36 settings is +0.073 (95\% $t$-CI [0.055, 0.092], $p=1.9\times10^{-9}$). The conclusion does not depend on the unit of analysis. Averaging first within benchmarks gives [0.023, 0.124] ($p=.014$); within models gives [0.059, 0.088] ($p=4.9\times10^{-5}$); and within the four families (Pythia, TinyLlama, SmolLM3, OLMo) gives +0.075, [0.048, 0.102] ($p=.003$). Averaged over models, the final gap is 0.070 on ARC-C, 0.068 on ARC-E, 0.078 on HellaSwag, 0.046 on MMLU, 0.161 on PIQA, and 0.017 on WinoGrande. Averaging the last two or three checkpoints instead of the final one gives similar results.

At the earliest checkpoints of Pythia and OLMo, the mean gap is $-0.008$ (95\% CI [$-0.018$, 0.003]); the low group is ahead in 13 of 24 settings (binomial $p=.84$). To measure how the groups diverge, we fit, for each setting, an ordinary least-squares slope of accuracy on $\log_{10}$ training tokens over the second half of the checkpoints, ordered by training step. The high-group slope exceeds the low-group slope by 0.020 per tenfold increase in tokens (positive in 26/36 settings; 95\% CI [0.011, 0.029]). The difference is positive for all six benchmark means (CI [0.010, 0.030]) and all six model means (CI [0.001, 0.039]); across the four family means it is +0.016 ($p=.13$).

\section{Controls}
\label{app:confounds}
\label{app:ablations}

\subsection{Mean Difficulty}
\label{app:mean-control}

Within length bins, high-TVS items have lower mean NLL in every benchmark. The within-bin Spearman correlations between \tvs and mean NLL are $-0.08$ (ARC-C), $-0.05$ (ARC-E), $-0.24$ (HellaSwag), $-0.10$ (MMLU), $-0.24$ (PIQA), and $-0.10$ (WinoGrande). The corresponding high-minus-low differences in mean NLL are $-0.03$, $-0.02$, $-0.19$, $-0.11$, $-0.32$, and $-0.13$ nats.

Because mean NLL strongly predicts accuracy, we also split on the residual of \tvs after regressing it on mean NLL within each benchmark and length bin. This difficulty-conditioned split roughly halves the final gap, from +0.073 (34/36) to +0.037 (26/36). What remains is concentrated in ARC-C (0.075), ARC-E (0.076), MMLU (0.056), and PIQA (0.050), with none on HellaSwag ($-0.014$) or WinoGrande ($-0.023$). Additionally conditioning on the PIT-4B gold-versus-distractor margin and PIT-4B accuracy leaves +0.014 (22/36).

The choice of dispersion statistic also matters. The unnormalized standard deviation of NLL reverses the pattern ($-0.026$, 8/36): it increases with mean NLL (within-bin Spearman $\rho=0.26$) and therefore selects harder items. A rank-based measure of disagreement, which is nearly uncorrelated with mean NLL, gives a weak positive gap (+0.011, 22/36).

\begin{table*}[!t]
  \centering
  \footnotesize
  \begin{tabular}{@{}llcc@{}}
    \toprule
    Signal & Control & Pearson & Spearman \\
    \midrule
    $\cv_{\mathrm{PPL}}$ & log mean PPL & 0.486 & 0.426 \\
    $\cv_{\mathrm{PPL}}$ & mean NLL & 0.443 & 0.423 \\
    $\cv_{\mathrm{PPL}}$ & length & $-$0.169 & $-$0.331 \\
    $\cv_{\mathrm{PPL}}$ & log length & $-$0.223 & $-$0.406 \\
    $\mathrm{TVS}$ & log mean PPL & 0.166 & 0.011 \\
    $\mathrm{TVS}$ & mean NLL & 0.134 & 0.007 \\
    $\mathrm{TVS}$ & length & $-$0.119 & $-$0.159 \\
    $\mathrm{TVS}$ & log length & $-$0.142 & $-$0.205 \\
    Length-controlled $\mathrm{TVS}$ & length & $-$0.016 & 0.006 \\
    Length-controlled $\mathrm{TVS}$ & log length & $-$0.011 & $-$0.004 \\
    Length-controlled $\mathrm{TVS}$ & mean PPL & 0.011 & 0.014 \\
    Length-controlled $\mathrm{TVS}$ & mean NLL & $-$0.079 & $-$0.091 \\
    \bottomrule
  \end{tabular}
  \caption{Correlations (with benchmark fixed effects) between the two dispersion measures (TVS and perplexity-based CV) and simple item properties over all 16,239 items. Length is the PIT-4B token length of the gold string. $\cv_{\mathrm{PPL}}$ is strongly entangled with mean difficulty and length. $\mathrm{TVS}$ is much less related to mean difficulty but still correlates with length, which motivates length-stratified grouping; after length control, correlations with length and mean PPL are near zero.}
  \label{tab:app-correlation-controls}
\end{table*}

\begin{figure*}[t]
  \centering
  \includegraphics[width=\textwidth,height=.74\textheight,keepaspectratio]{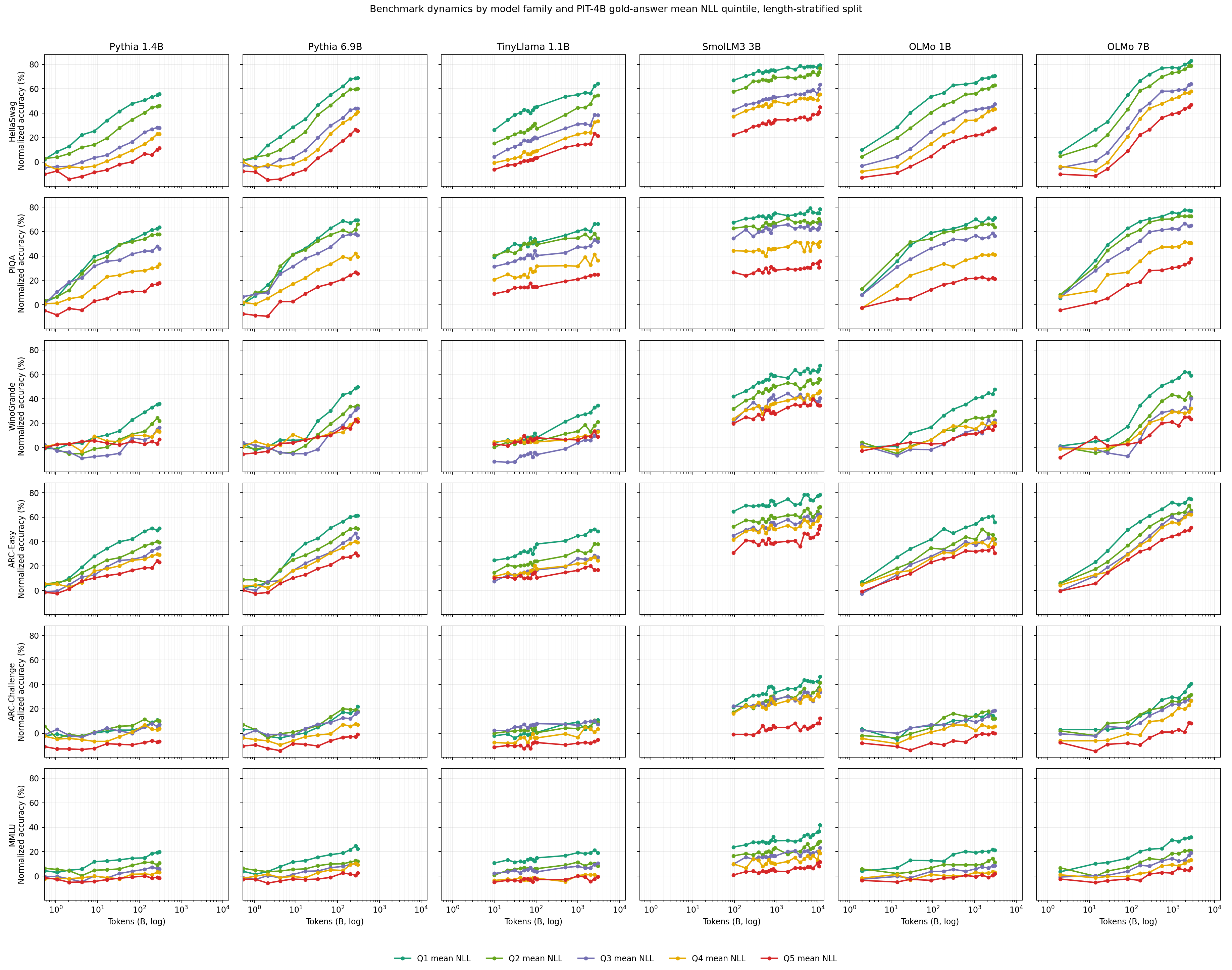}
  \caption{Split by mean gold NLL. Higher mean NLL is associated with lower accuracy at all checkpoints. Difficulty-conditioned \tvs results are reported separately in Appendix~\ref{app:mean-control}.}
  \label{fig:app-mean-nll}
\end{figure*}

\subsection{Length}
\label{app:length-control}

Raw \tvs decreases with string length, so we form all groups within length bins. For each benchmark, we sort items by the PIT-4B token length of the gold string, divide them into ten bins, rank items by \tvs within each bin, and assign low, medium, and high groups within the bin before pooling. After this stratification, length-controlled \tvs is essentially uncorrelated with length (Pearson $-0.016$, Spearman $0.006$; Table~\ref{tab:app-correlation-controls}), so the high-TVS advantage is not explained by high-TVS items being shorter or longer.

\subsection{Perplexity-Based CV}
\label{app:ppl-cv-bad}

An alternative is the coefficient of variation of perplexity rather than that of NLL. Because perplexity exponentiates NLL, its dispersion grows with its level: $\cv_{\mathrm{PPL}}$ correlates at about 0.43--0.49 with mean perplexity and mean NLL (Table~\ref{tab:app-correlation-controls}), which makes it hard to separate temporal variation from difficulty. We therefore use \tvs throughout.

\section{Alternative Target Strings}
\label{app:alt-targets}

We compute \tvs on the question together with its gold answer, because the question alone often underspecifies the relevant fact or procedure. Two alternative targets test this choice.

\subsection{Question-Only TVS}
\label{app:question-only}

\begin{figure*}[t]
  \centering
  \includegraphics[width=\textwidth,height=.74\textheight,keepaspectratio]{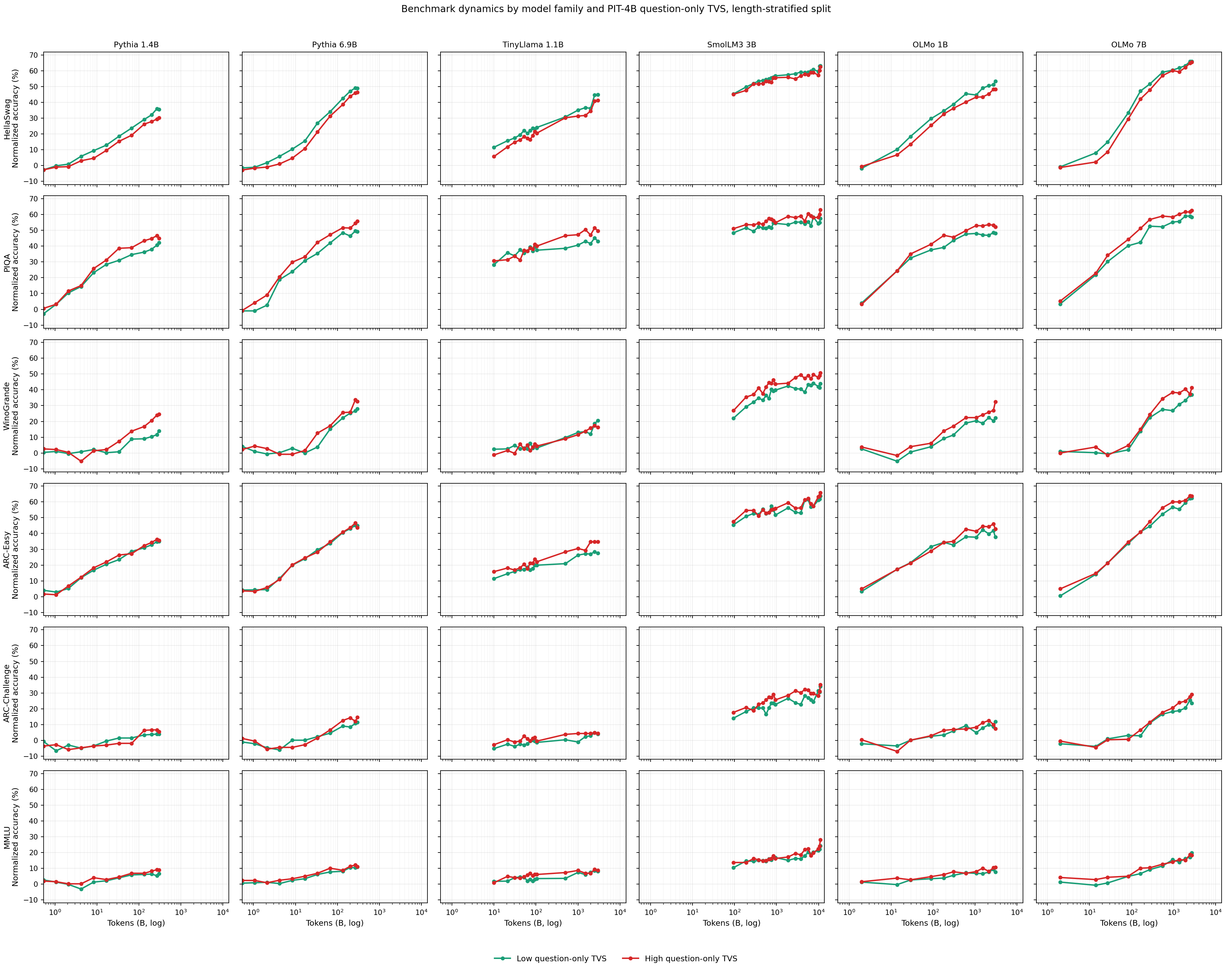}
  \caption{Split by question-only \tvs. The effect is much weaker than for the question and gold answer together, so the signal is not reducible to the question stem.}
  \label{fig:app-question-only}
\end{figure*}

Computing \tvs on the question alone yields a much weaker effect: high-TVS items outperform low-TVS items in 26 of 36 final settings, by 0.021 on average, compared with 0.073 for the question and gold answer together (Figure~\ref{fig:app-question-only}). The signal therefore depends substantially on the correct answer and not only on the wording of the question.

\subsection{Gold Answer versus Best Wrong Answer}
\label{app:gold-wrong}

\begin{figure*}[t]
  \centering
  \includegraphics[width=\textwidth,height=.74\textheight,keepaspectratio]{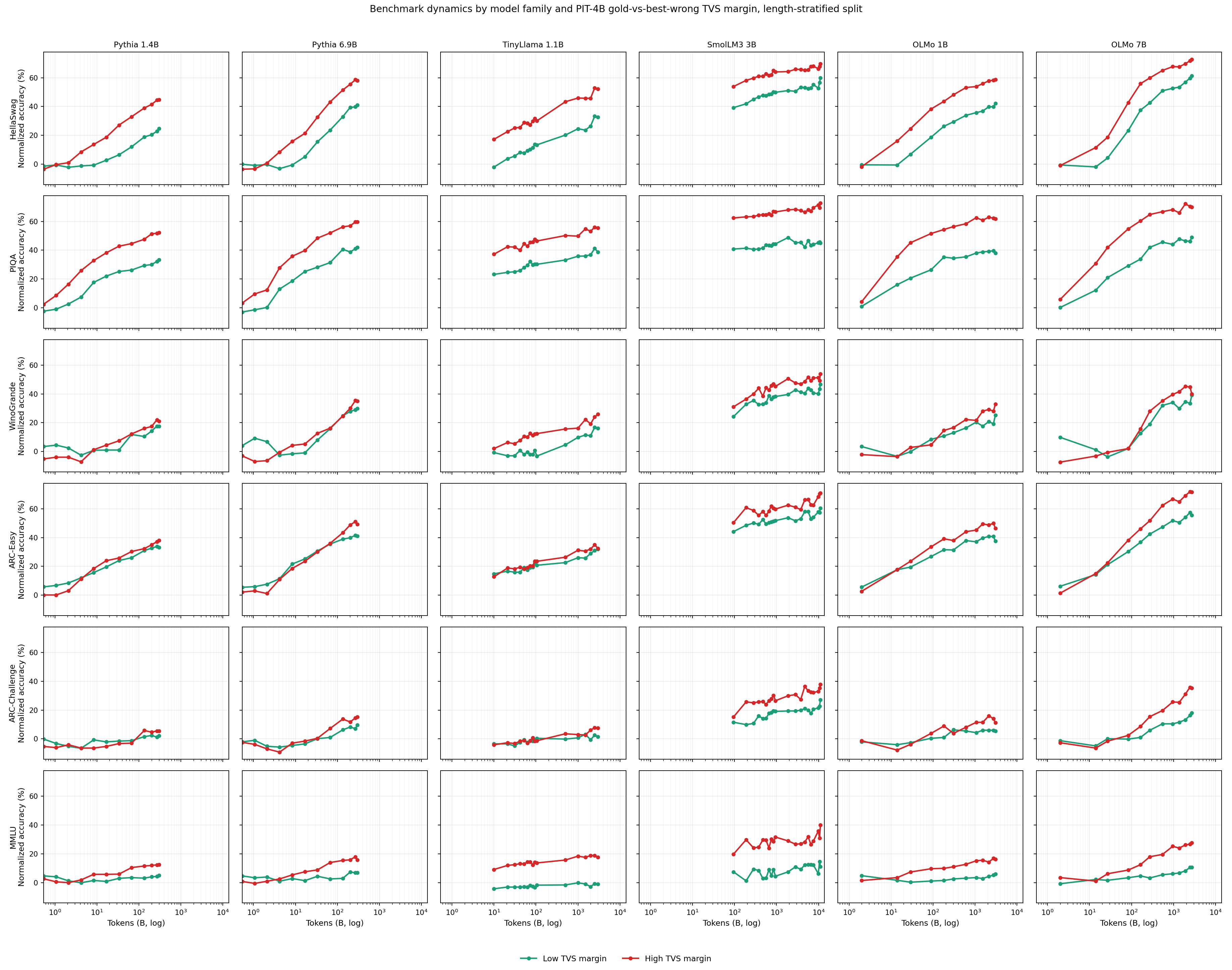}
  \caption{Split by the gold-versus-wrong \tvs margin. Items whose gold answer is more temporally variable than the best wrong answer show the largest final-checkpoint advantage.}
  \label{fig:app-gold-wrong}
\end{figure*}

To isolate variability specific to the correct answer, we subtract the \tvs of the best wrong answer from that of the gold answer:
\begin{equation}
  \begin{aligned}
    \Delta \mathrm{TVS}(i)
    &=
    \mathrm{TVS}(q_i, a_i^{\mathrm{gold}}) \\
    &\quad -
    \mathrm{TVS}(q_i, a_i^{\mathrm{wrong}}).
  \end{aligned}
\end{equation}
where $\mathrm{TVS}(q,a)$ is \tvs computed on question $q$ followed by answer $a$. Splitting on this margin gives the largest final-checkpoint separation of any stratification. The high-margin group is ahead in all 36 final settings and is the best group in 35 of them, with a mean gap of 0.123 (Table~\ref{tab:app-split-summary}, Figure~\ref{fig:app-gold-wrong}). At the earliest Pythia and OLMo checkpoints, the low-margin group is ahead in 18 of 24 settings, so this advantage emerges during training.

\section{Item-Level Analysis by Benchmark}
\label{app:item-level}

We decompose \tvs into a component shared with the distractors (the mean \tvs over wrong-answer strings) and a gold-specific component (gold minus distractor \tvs). For each benchmark, we regress two item-level outcomes on both components. The first is correctness at the final checkpoint. The second is late-training margin growth: the least-squares slope of the evaluated model's gold-versus-best-distractor margin (in mean NLL) on $\log_{10}$ tokens, over checkpoints in the upper half of the log-token range. Each regression controls for PIT-4B mean gold NLL, the PIT-4B gold-versus-best-distractor margin, PIT-4B accuracy, and fixed effects for length bin and model, with standard errors clustered by item. Table~\ref{tab:app-item-level} reports effects per standard deviation of the gold-specific component.

The gold-specific component strongly predicts late margin growth on ARC-C, ARC-E, and MMLU, weakly on PIQA, and not at all on HellaSwag or WinoGrande, where it instead predicts a small difference in final correctness. The benchmarks also differ in how \tvs is composed. On the three benchmarks posed as ``Question: \ldots{} Answer:'' with short answers (ARC-C, ARC-E, MMLU), 90--92\% of the variance in gold \tvs is shared with the distractors, compared with 53--59\% on HellaSwag and WinoGrande.

\section{What Does TVS Capture?}
\label{app:construct}

We report two further analyses of what \tvs captures: a controlled synthetic intervention and a comparison with training-data frequency.

\paragraph{Synthetic intervention.}
For each of five seeds, we trained eight small causal language models, one per synthetic ``era.'' Token, entity, and answer counts and total support were held fixed across eras; only the allocation of each answer's conditional support over eras differed. Making support era-dependent raised \tvs from 0.229 to 0.351 (5/5 seeds; paired $t(4)=11.2$, $p=3.6\times10^{-4}$), whereas a stable 50/50 mixture of conflicting evidence did not raise it. Models were trained for 6 epochs; a single-seed pilot trained for 12 epochs showed the same direction (+0.156, $p=.09$). The experiment shows that \tvs responds to era-dependent conditional support when other corpus statistics are held fixed.

\paragraph{Training-data frequency.}
Using 25.7B tokens sampled from the training mixture of OLMo-7B-0724, we counted 2- and 3-grams ending in each gold answer and the corresponding gold-versus-distractor frequency margins. We then regressed the change in OLMo-7B-0724's gold margin between 83B and 2457B training tokens on these counts and \tvs. Low answer frequency is the dominant predictor (partial $R^2\approx0.021$). Controlling for these counts, \tvs retains an independent association ($\beta=0.031$ per SD, 95\% CI [0.017, 0.045], HC3 standard errors; $\Delta R^2\approx0.001$), concentrated in ARC-C, ARC-E, and MMLU ($I^2=74\%$), in line with the item-level results in \S\ref{sec:where}. Because $n$-gram counts capture surface wording, this association is consistent with \tvs reflecting variation beyond exact textual matches.

\section{Contamination Checks}
\label{app:contamination}

\paragraph{Release-date test.}
If copies of benchmark items entered later crawls, PIT-4B models trained on crawls from after a benchmark's release should assign its gold answers lower NLL than the year trend shared by all benchmarks would predict. We take the release years to be 2018 for ARC, 2019 for HellaSwag and WinoGrande, 2020 for PIQA, and 2021 for MMLU. We estimate a staggered difference-in-differences model with item and year fixed effects. Post-release gold NLL is slightly \emph{higher} (+0.016), with no detectable gold-versus-distractor margin change ($p=.34$), and PIT-4B accuracy on high-TVS items falls ($-0.011$, $p=.004$). ARC, PIQA, and MMLU also share the same year profile despite their different release years. This test finds no release-specific improvement predicted by benchmark leakage into PIT-4B's training snapshots.

\paragraph{Verbatim overlap.}
Following \citet{brown-etal-2020-language}, we scanned 55.4B tokens of FineWeb, ten files per crawl year from 2017 to 2024, for 13-gram overlaps with each item. We searched separately for the question alone, for 13-grams spanning the question and the gold answer, and for the gold answer alone, using the same spans for wrong answers as controls. Overlap is rare and mostly comes from source text, such as WikiHow pages, rather than from pages that pair questions with their answers. Any overlap occurs for 3.1\% of high-TVS and 2.0\% of low-TVS items (length-stratified OR $=1.56$, $p=3\times10^{-4}$). For gold-answer text the figures are 2.1\% versus 1.0\% (OR $=1.90$), and for distractor text 0.6\% versus 0.3\% (OR $=1.56$). Within items, years with more overlap do not have lower PIT-4B NLL (question: $p=.63$; question--gold span: $p=.60$; gold answer: $p=.08$). Adding overlap indicators to the item-level regressions leaves the \tvs coefficient essentially unchanged (0.0267 versus 0.0268 for final correctness). These checks provide evidence against verbatim overlap as the dominant explanation of the \tvs pattern.

\end{document}